\documentclass[letterpaper, 10 pt, conference]{ieeeconf}  

\IEEEoverridecommandlockouts                              

\usepackage{standalone}
\usepackage[textsize=tiny]{todonotes}
\usepackage{tabularx}
\usepackage{pgfplots, pgfplotstable}
\usepackage{textcomp}
\usepackage{makecell}
\usepackage{booktabs}
\usepackage{listings}
\usepackage{lipsum}
\usepackage{multirow}
\usepackage{color}
\usepackage[T1]{fontenc}
\usepackage{amsmath, amssymb}
\usepackage{graphicx}
\usepackage[labelfont=bf]{caption}
\usepackage{bm}
\usepackage{afterpage}

\usepackage{amsmath}        
\usepackage[ruled, vlined, linesnumbered]{algorithm2e}

\usepackage{float}

\makeatletter
\newcommand\BeraMonottfamily{%
  \def\fvm@Scale{0.85}
  \fontfamily{fvm}\selectfont
}
\makeatother

\usepackage{xcolor}
\usepackage{array}
\usepackage{wasysym}
\usepackage[flushleft]{threeparttable}

\usepackage[T1]{fontenc}
\usepackage{hyperref}
\usepackage{preamble}

\usepackage{cleveref}
\usepackage{graphicx}
\usepackage{booktabs}
\usepackage{pifont}

\title{\LARGE \bf

TAME-Nav: Temporal Asymmetric Mixture-of-Experts Distillation for Agile and Versatile Robot Navigation

}

\author{
Lingjie Zhang$^{1,2}$,
Jun Xue$^{1,2}$,
Shanze Wang$^{2,3}$,
Haoyu Xi$^{1,2}$,
Xinming Zhang$^{2,4}$,
Xiaoyu Shen$^{2}$,
Wei Zhang$^{2,*}$
}

\begin{document}

\newcommand{\jh}[1]{{ \color{blue}[jh: #1]}}

\maketitle

\begingroup
\renewcommand{\thefootnote}{}
\footnotetext[0]{%
\begin{tabular}{@{}r@{\hspace{0.4em}}p{\dimexpr\columnwidth-1.8em\relax}@{}}
$^{1}$ & Shanghai Jiao Tong University, Shanghai, China.\\
$^{2}$ & Eastern Institute of Technology, Ningbo, China.\\
$^{3}$ & Department of Aeronautical and Aviation Engineering,
          The Hong Kong Polytechnic University, Hung Hom, Kowloon, Hong Kong.\\
$^{4}$ & University of Science and Technology of China, Hefei, China.\\
$^{*}$ & Wei Zhang is the corresponding author: zhw@eitech.edu.cn. \\ \\
$^{}$ & For videos and code, see \href{https://tame-nav.github.io/}{tame-nav.github.io}.
\end{tabular}%
}
\endgroup

\thispagestyle{empty}
\pagestyle{empty}

\begin{abstract}

Agile and reliable navigation is a core capability for embodied robots navigating complex spatial constraints and dynamic interactions. However, a single policy often struggles to accommodate the distinct decision-making demands of diverse scenarios. We present TAME-Nav, an asymmetric Teacher–Student navigation framework built around a temporal soft mixture of experts (Soft-MoE). The Teacher combines clean observations with causally maintained dynamic-obstacle tracks to construct privileged representations, while the Student learns complementary navigation features from corrupted observation histories through temporal encoding, soft expert routing, and latent distillation. Procedural scenario generation and parallel task-level simulation broaden training coverage, while deployment-informed asymmetric domain randomization supports sim-to-real transfer. Across in-distribution and unseen compositional scenarios, TAME-Nav achieves higher success rates, lower collision rates, and higher navigation scores than representative baselines. On a real quadruped robot, it completes all twenty trials across local-trap escape and narrow-passage navigation with sudden dynamic obstacles, with only one minor contact.

\end{abstract}

\section{Introduction}

Reliable local navigation is a prerequisite for autonomous navigation in complex environments. A local navigation policy converts onboard sensory observations and a local goal into real-time motion commands, determining whether a planned route can be executed safely and continuously~\cite{hoeller2021learning}. A capable local navigator can also reduce the burden on upstream modules by allowing them to provide sparser subgoals at a lower decision frequency~\cite{cheng2024navila}. Traditional model-based planners offer interpretable motion models and geometric constraints, but their performance often depends on manually designed objectives and short-horizon environment predictions~\cite{fox1997dynamic}. Consequently, they may become trapped near concave obstacles, oscillate in narrow passages~\cite{koren1991potential}, or exhibit overly conservative freezing behavior in dense, interacting crowds~\cite{trautman2010unfreezing}. Learning-based methods can acquire richer interaction priors from demonstrations or environmental experience~\cite{pfeiffer2017perception}, yet their performance remains strongly affected by the diversity of the training distribution and the expressiveness of the observation representation~\cite{xu2022learning}. Learned policies may generalize poorly to unseen cluttered or confined environments~\cite{zhang2026saferpath}. Obstacle motion must be inferred from sensory histories, while explicit object-state inputs introduce deployment-time tracking dependencies~\cite{xu2025navrl}.

These raise two fundamental questions: 

\begin{enumerate}
\item[\textbf{Q1.}] With only local observations, how can a single policy coordinate complementary navigation skills, including safe obstacle avoidance, narrow-passage traversal, and local-trap escape?
\item[\textbf{Q2.}] Can such coordination translate into higher task completion rates and lower collision risk in unseen compositional scenarios and real-world environments?
\end{enumerate}

To address these questions, we present TAME-Nav, an asymmetric
Teacher--Student framework for agile and versatile robot navigation.
The Teacher fuses clean observation histories with causal dynamic-object
tracks, while a temporal Soft-MoE Student learns to match its privileged
navigation representation through latent distillation. Both branches
share an Actor that generates body-frame velocity commands. Using only
onboard observations and local goals, the deployed Student supports
obstacle avoidance, narrow-passage traversal, and local-trap recovery
without explicit object tracking or global-map inputs.

We further develop a massively parallel task-level simulator with
procedurally generated geometric layouts and dynamic interactions to
broaden training coverage. Deployment-informed randomization models
sensor degradation and velocity-response uncertainty: the Student
receives corrupted observations, while the Teacher and Critic retain
clean observations of the underlying simulated state.

Our evaluations demonstrate consistent gains in navigation success
and safety over SEA-Nav, NavRL, and DRL-DCLP under matched training
budgets. TAME-Nav improves in-distribution success by over 25 percentage
points over the strongest baseline and maintains its advantage across
unseen sequential and multiroom scenarios. Real-world experiments
further demonstrate its ability to escape local traps and negotiate
narrow passages with suddenly appearing obstacles: the quadruped
completes all trials in both scenarios, with only one minor contact
across the twenty trials.

In summary, our main contributions are as follows.
\begin{enumerate}
    \item We propose TAME-Nav, an asymmetric Teacher--Student framework that transfers privileged navigation representations augmented with causal dynamic-obstacle tracks from a Teacher to a deployable temporal Soft-MoE Student, enabling a single policy to handle obstacle avoidance, constrained-space traversal, and local-trap escape.

    \item We develop a massively parallel task-level simulator with procedural scene generation for diverse geometric and dynamic navigation challenges, together with deployment-informed asymmetric randomization that corrupts Student observations while preserving clean Teacher--Critic supervision.

    \item We demonstrate improved navigation success and safety across
    in-distribution and unseen compositional scenarios, and validate
    reliable local-trap escape and narrow-passage navigation with
    sudden dynamic obstacles on a real quadruped.
\end{enumerate}

\section{Related Work}

\subsection{Learning-Based Local Navigation}

Local navigation methods can be broadly categorized into model-based approaches and learning-based approaches. Classical methods such as DWA and TEB search for safe control commands online using explicit kinematic constraints and manually designed cost functions~\cite{fox1997dynamic, rosmann2013efficient}. Although these methods are interpretable and computationally efficient, their performance in complex static–dynamic environments often depends heavily on environment modeling and parameter tuning. Reinforcement learning instead learns a direct mapping from observations, such as LiDAR measurements and goal states, to continuous velocity commands~\cite{tai2017virtual}. Subsequent studies have incorporated dynamic-agent modeling, velocity-obstacle priors, and temporal environment representations to improve anticipatory collision avoidance in crowded environments~\cite{xie2023drl, hoeller2021learning}. Recent work has also explored learning-based locomotion and navigation for quadruped robots in high-speed, cluttered, and open environments~\cite{he2024agile, ren2024top}. Nevertheless, existing methods are often limited by insufficient coverage of diverse local-navigation scenarios and the high cost of data collection in high-fidelity physics simulators.

\subsection{Teacher–Student Reinforcement Learning}

Teacher–Student reinforcement learning typically allows the Teacher to access privileged information during training and transfers the acquired capability to a Student that relies only on deployable observations, commonly through action imitation or latent-representation distillation~\cite{pinto2017asymmetric, chen2020learning}. Recent studies have further shown that excessive reliance on privileged information unavailable to the Student may lead the Teacher to develop behaviors that are difficult to imitate, highlighting the need to account for the Teacher–Student information gap during Teacher training~\cite{messikommer2025student}.

This paradigm has been widely adopted in legged locomotion, where privileged terrain information or full-state observations are used to train a Teacher, while the deployable Student operates using proprioceptive signals and observation histories~\cite{lee2020learning, miki2022learning}. Wu et al.~\cite{wu2026toward} further incorporate a gated MoE Student into concurrent Teacher--Student learning to improve
proprioceptive locomotion across diverse terrains. In navigation, DiPCAN reconstructs privileged dynamic-environment representations from historical visual observations~\cite{monaci2022dipcan}, while the recent GPTS-Nav employs graph-structured privileged information to guide a LiDAR-based navigation policy in dynamic environments~\cite{zheng2026gpts}. These studies demonstrate the value of privileged knowledge for action generation; however, most focus on observation asymmetry within the same simulation setting and employ a unified Student representation for a relatively specific class of navigation scenarios.

Our approach exploits a lightweight task-level simulator to achieve broader coverage of complex maps and failure cases. The Teacher–Student framework transfers privileged navigation knowledge to a deployable Student, while the Student’s mixture-of-experts (MoE) architecture provides a structural basis for learning scene-dependent representations across heterogeneous local-navigation situations.

\section{Method}

\begin{figure*}[!t]
    \vspace{5pt}
    \centering
    \includegraphics[width=\textwidth]{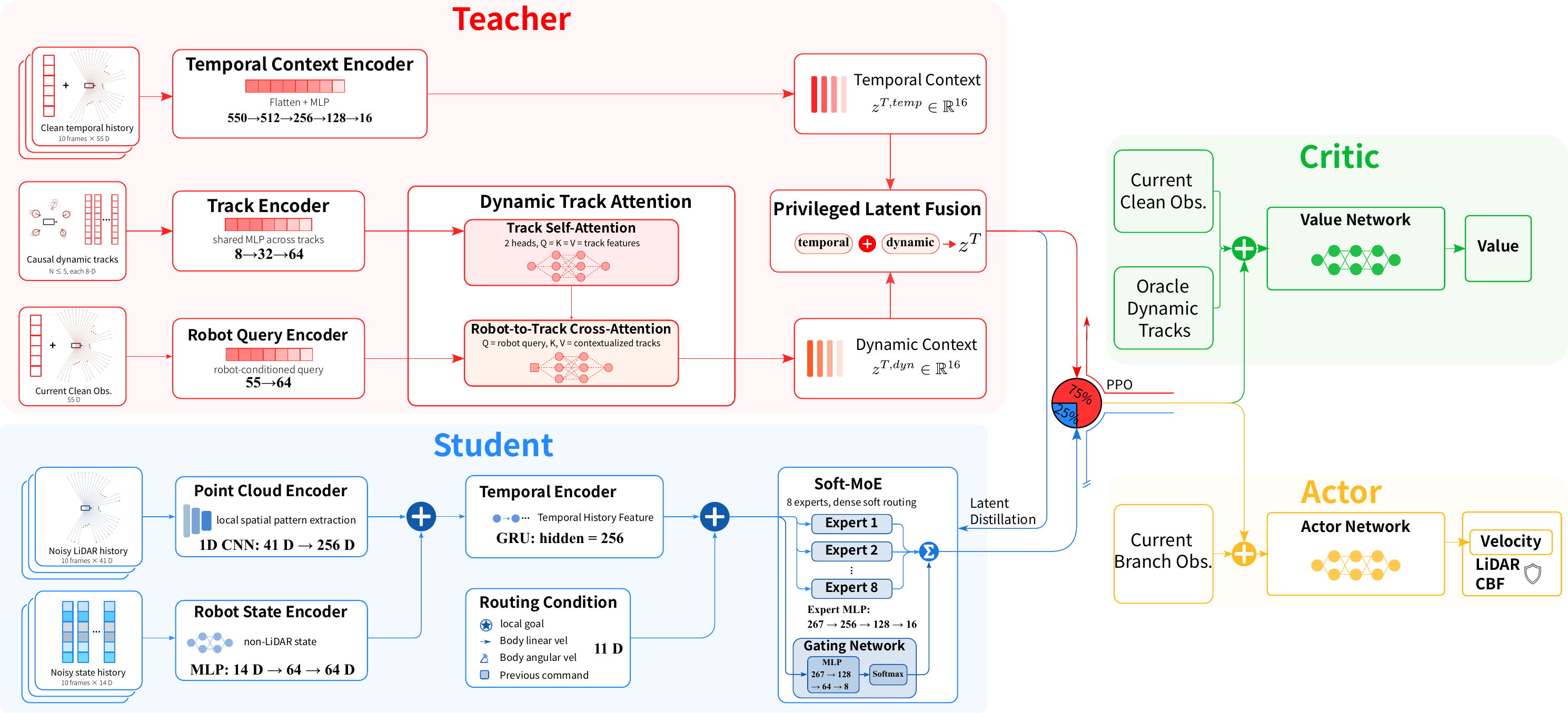}
    \caption{ Overview of TAME-Nav. The Teacher combines clean observation histories with causal dynamic tracks, while the temporal Soft-MoE Student learns from corrupted histories through latent distillation. Both share an Actor with LiDAR-based safety correction. A privileged Critic supports training; only the Student encoder, Actor, and safety module are deployed.}
    \label{fig:framework}
    \vspace{-12pt}
\end{figure*}


Coordinating diverse navigation behaviors within a single deployable policy remains challenging. We propose TAME-Nav, an asymmetric Teacher–Student framework centered on a temporal Soft-MoE Student. The following subsections present the navigation formulation, network architecture, procedural training environments, and asymmetric domain randomization.

\subsection{Preliminaries}

\subsubsection{Local Navigation Formulation}
\label{subsec:local_navigation_formulation}

We formulate navigation in dynamic environments as a partially
observable Markov decision process
(POMDP)~\cite{kaelbling1998planning}.
Scenarios $\xi\sim p(\xi)$ vary in obstacle geometry,
start--goal configurations, and dynamic interactions.
The robot receives LiDAR measurements, the relative goal,
its motion state, and the previous command, as defined in
Eq.~\eqref{eq:local_observation}:
\begin{equation}
o_t=
\left[
o_t^{\mathrm{lidar}},
o_t^{\mathrm{goal}},
o_t^{\mathrm{motion}},
\mathbf u_{t-1}
\right].
\label{eq:local_observation}
\end{equation}

To account for motion and occlusion, the policy conditions on
an $H$-frame history $h_t=(o_{t-H+1},\ldots,o_t)$ and generates
$\mathbf u_t\sim\pi_\theta(\cdot\mid h_t)$.
The command $\mathbf u_t=[v_{x,t},v_{y,t},\omega_{z,t}]^\top$
specifies body-frame longitudinal velocity, lateral velocity,
and yaw rate.

We learn safe and efficient goal-reaching behavior by maximizing
the expected discounted return in Eq.~\eqref{eq:policy_objective}:
\begin{equation}
J(\pi_\theta)=
\mathbb E_{\xi,\tau}
\left[\sum_{t=0}^{T-1}\gamma^t r_t\right].
\label{eq:policy_objective}
\end{equation}
Here, $\tau$ is a policy-generated trajectory in scenario $\xi$,
$r_t$ is the reward, and $\gamma\in[0,1)$ is the discount factor.
The episode ends at $T$ upon success, failure, or reaching
the maximum horizon.

\subsection{Overall Architecture}



As illustrated in Fig.~\ref{fig:framework}, TAME-Nav adopts an
asymmetric Teacher--Student architecture with separate encoders
and a shared high-level Actor. The Teacher combines clean observation
histories with causal dynamic-obstacle tracks, while the temporal
Soft-MoE Student infers a compatible representation solely from
onboard histories. During deployment-oriented training, Student
observations are delayed and corrupted, while the Teacher and
Critic retain clean observations and privileged information.

Given the current observation and selected latent, the Actor
produces a nominal body-frame velocity mean, which a differentiable
LiDAR-based CBF corrects to parameterize a Gaussian policy.
Teacher and Student collect rollouts from fixed subsets of parallel
environments, with 75\% allocated to Teacher-controlled rollouts and 25\% to Student-controlled rollouts. PPO~\cite{schulman2017proximal} updates the Teacher
encoder, shared Actor, and privileged Critic, while the Student
encoder learns through latent distillation from stop-gradient
Teacher representations at Student-visited states.
Deployment retains only the Student encoder, shared Actor,
and LiDAR-based safety module.

\textbf{Track-Augmented Teacher}:
The Teacher combines clean observation histories with causal
dynamic-obstacle tracks to construct a privileged representation
for the shared Actor and Student latent distillation.
Following Eq.~\eqref{eq:local_observation}, an MLP encodes a
ten-frame history $\bar{\mathbf H}_t^T\in\mathbb R^{10\times55}$
into
$\mathbf z_t^{T,\mathrm{temp}}
=E_T^{\mathrm{temp}}(\operatorname{vec}(\bar{\mathbf H}_t^T))
\in\mathbb R^{16}$.

The training environment maintains causal tracks and selects
the five nearest valid obstacles, forming
$\mathcal B_t=\{\mathbf b_t^i\}_{i=1}^{N_t}$ with $N_t\leq5$.
Each track is defined in Eq.~\eqref{eq:teacher_track}:
\begin{equation}
\mathbf b_t^i =
\left[
(\mathbf p_{t,\mathrm{rel}}^i)^\top,
(\mathbf v_{t,\mathrm{obs}}^{i,B})^\top,
\rho_t^i,
\sigma_t^i,
f_{t,\mathrm{miss}}^i,
m_{t,\mathrm{vis}}^i
\right]^\top
\in\mathbb R^8.
\label{eq:teacher_track}
\end{equation}
Here, $\mathbf p_{t,\mathrm{rel}}^i$ and
$\mathbf v_{t,\mathrm{obs}}^{i,B}$ denote robot-relative position
and obstacle velocity, both expressed in body coordinates.
The remaining entries represent circumscribed radius,
age-dependent heuristic uncertainty, normalized observation age,
and visibility. Occluded tracks are extrapolated using their
latest velocity estimates until expiration; no future obstacle
states are accessed.

Normalized track features pass through a shared MLP and two-head
self-attention with a residual connection and
LayerNorm~\cite{vaswani2017attention}.
The resulting features $\widetilde{\mathbf E}_t$ are padded to
five entries and aggregated by two-head cross-attention, using
$\mathbf q_t=\phi_{\mathrm{qry}}(\bar{\mathbf o}_t^T)
\in\mathbb R^{64}$ as the robot-conditioned query and
$\widetilde{\mathbf E}_t$ as keys and values.
Each head has dimension $32$, and padded entries are masked.
LayerNorm and a linear projection produce
$\mathbf z_t^{T,\mathrm{dyn}}\in\mathbb R^{16}$, which is set
to zero when no valid tracks exist.

The final latent uses normalized residual fusion:
\begin{equation}
\mathbf z_t^T =
\operatorname{Norm}_2
\left(
\operatorname{sg}(\mathbf z_t^{T,\mathrm{temp}})
+g_{\mathrm{dyn}}\mathbf z_t^{T,\mathrm{dyn}}
\right).
\label{eq:teacher_latent_fusion}
\end{equation}
In Eq.~\eqref{eq:teacher_latent_fusion},
$\operatorname{Norm}_2$ denotes L2 normalization, and
$g_{\mathrm{dyn}}$ is a learnable scalar initialized to one.
Stop-gradient blocks Actor gradients to the temporal encoder,
which learns through the value loss; the dynamic encoder and
fusion coefficient receive policy gradients.

\textbf{Temporal Soft-MoE Student}: The Student combines temporal encoding with softly routed expert representations to
construct a navigation latent from onboard sensing and control
histories. Its observation $\widetilde{\mathbf o}_t^S\in\mathbb R^{55}$ comprises a
41-dimensional LiDAR scan and 14 non-LiDAR quantities:
the body-frame goal, linear and angular velocities, previous command, and projected gravity. The tilde denotes
observations potentially affected by sensing delays and corruption.

The Student maintains a history
$\widetilde{\mathbf H}_t^S\in\mathbb R^{10\times55}$.
The window advances with the $10\,\mathrm{Hz}$ LiDAR stream,
covering a nominal $0.9\,\mathrm{s}$ scan history.
Between updates, the latest scan and associated goal observation
are held, while the newest slot's proprioceptive state and previous
command are refreshed at the $50\,\mathrm{Hz}$ control rate.

For each history slot, a three-layer 1D CNN encodes LiDAR into
$\mathbf f_k^{\mathrm{lidar}}\in\mathbb R^{256}$, and a two-layer
MLP encodes the remaining state into
$\mathbf f_k^{\mathrm{state}}\in\mathbb R^{64}$.
Their concatenation
$\mathbf f_k=[\mathbf f_k^{\mathrm{lidar}};
\mathbf f_k^{\mathrm{state}}]\in\mathbb R^{320}$
is processed by a single-layer GRU~\cite{cho2014learning}.
Its final hidden state summarizes the observation sequence
in Eq.~\eqref{eq:student_temporal}:
\begin{equation}
\mathbf h_t^S
=
\operatorname{GRU}
\left(\mathbf f_1,\ldots,\mathbf f_{10}\right)
\in\mathbb R^{256}.
\label{eq:student_temporal}
\end{equation}

An eight-expert Soft-MoE conditions this summary on the
11-dimensional navigation context $\boldsymbol\chi_t^S$,
comprising the latest goal, linear and angular velocities,
and previous command.
Let $\mathbf x_t^S=[\mathbf h_t^S;\boldsymbol\chi_t^S]
\in\mathbb R^{267}$.
The expert branches share an input projection and produce
separate 16-dimensional candidate representations.
A learned router combines them through the dense soft
aggregation~\cite{jacobs1991adaptive} in Eq.~\eqref{eq:student_soft_moe}:
\begin{equation}
\begin{aligned}
\mathbf w_t
&=
\operatorname{softmax}\left(G_S(\mathbf x_t^S)\right),\\
\mathbf z_t^S
&=
\operatorname{Norm}_2
\left(
\sum_{i=1}^{8}w_t^i E_i(\mathbf x_t^S)
\right)
\in\mathbb R^{16}.
\end{aligned}
\label{eq:student_soft_moe}
\end{equation}
Here, $E_i$ denotes the $i$-th expert mapping and
$\operatorname{Norm}_2$ denotes L2 normalization.
This formulation supports context-dependent combinations of
expert features through continuous, differentiable routing.

The Student is trained by matching the Teacher's fused
representation on Student-controlled rollout samples.
For a minibatch $\mathcal D_S$, its objective is given in
Eq.~\eqref{eq:student_distillation}:
\begin{equation}
\mathcal L_S
=
\frac{1}{16|\mathcal D_S|}
\sum_{t\in\mathcal D_S}
\left\|
\mathbf z_t^S-\operatorname{sg}(\mathbf z_t^T)
\right\|_2^2
+
\lambda_{\mathrm{bal}}\mathcal L_{\mathrm{bal}}.
\label{eq:student_distillation}
\end{equation}
The Teacher targets correspond to the same rollout states
and are held fixed during Student optimization.
The balancing term penalizes deviations of minibatch-mean
expert weights from a uniform distribution.
Only the Student encoder receives gradients from this
objective; PPO gradients do not update it.
The resulting representation replaces the Teacher latent
at the shared Actor interface during deployment.


\textbf{Shared Actor and Privileged Critic}: The Teacher and Student share a high-level Actor conditioned on
the current observation and the corresponding navigation latent.
For branch $b\in\{T,S\}$, the shared backbone produces
$\mathbf h_t^b=\phi_\pi([\mathbf o_t^b;\mathbf z_t^b])$.
Two heads predict the nominal body-frame velocity
$\boldsymbol{\mu}_t^{b,\mathrm{nom}}
=\phi_{\mathrm{nav}}(\mathbf h_t^b)$ and a positive CBF
coefficient
$\alpha_t^b=\operatorname{softplus}(\phi_\alpha(\mathbf h_t^b))$.

We adopt SEA-Nav's differentiable LiDAR-based CBF
formulation~\cite{chen2026sea}, which corrects the translational
velocity mean while preserving the nominal yaw rate.
The corrected mean parameterizes the Gaussian policy in
Eq.~\eqref{eq:shared_actor_policy}:
\begin{equation}
\begin{aligned}
\boldsymbol{\mu}_t^{b,\mathrm{cbf}}
&=
\mathcal P_{\mathrm{CBF}}
\left(
\boldsymbol{\mu}_t^{b,\mathrm{nom}};
\mathbf o_t^{b,\mathrm{lidar}},\alpha_t^b
\right),\\
\pi_\theta(\mathbf u_t\mid\mathbf o_t^b,\mathbf z_t^b)
&=
\mathcal N
\left(
\boldsymbol{\mu}_t^{b,\mathrm{cbf}},
\operatorname{diag}(\boldsymbol{\sigma}^2)
\right),
\end{aligned}
\label{eq:shared_actor_policy}
\end{equation}
where $\boldsymbol{\sigma}\in\mathbb R^3$ is a learnable,
state-independent standard deviation.

The training-only Critic receives additional privileged
information~\cite{pinto2017asymmetric}. An obstacle-set encoder maps the ground-truth
states of up to five dynamic obstacles---body-frame relative
positions, obstacle velocities, and radii---to
$\mathbf d_t^V
=E_V^{\mathrm{dyn}}(\mathcal S_t^{\mathrm{dyn}},
\bar{\mathbf o}_t^T)\in\mathbb R^{32}$.
A normalized static cost-to-go scalar $c_t^{\mathrm{stat}}$
is queried from a goal-conditioned Soft-min field.
Together with the current Teacher observation and its temporal
latent, these features form the 104-dimensional value input
in Eq.~\eqref{eq:privileged_critic}:
\begin{equation}
V_\psi(s_t)
=
\phi_V
\left(
[
\bar{\mathbf o}_t^T;
c_t^{\mathrm{stat}};
\mathbf z_t^{T,\mathrm{temp}};
\mathbf d_t^V
]
\right).
\label{eq:privileged_critic}
\end{equation}
This privileged information supports PPO value estimation
without entering the Actor. Deployment retains only the
Student encoder, shared Actor, and LiDAR-based safety module.

\subsection{Task-Level Simulation and Procedural Scenario Generation}

We train TAME-Nav in a custom GPU-vectorized 2D navigation simulator.
A command-response model updates robot poses and velocities without
simulating joint-level dynamics. The simulator retains footprint
collisions with blocking and sliding, velocity and acceleration
limits, dynamic obstacles, and LiDAR and odometry observations.
At deployment, a separately trained locomotion controller executes
the same velocity commands. This abstraction supports large-scale
interaction and flexible scenario generation, with sensing and
execution perturbations introduced during deployment-oriented training.

\begin{figure}[t]
    \vspace{5pt}
    \centering
    \includegraphics[width=\columnwidth]{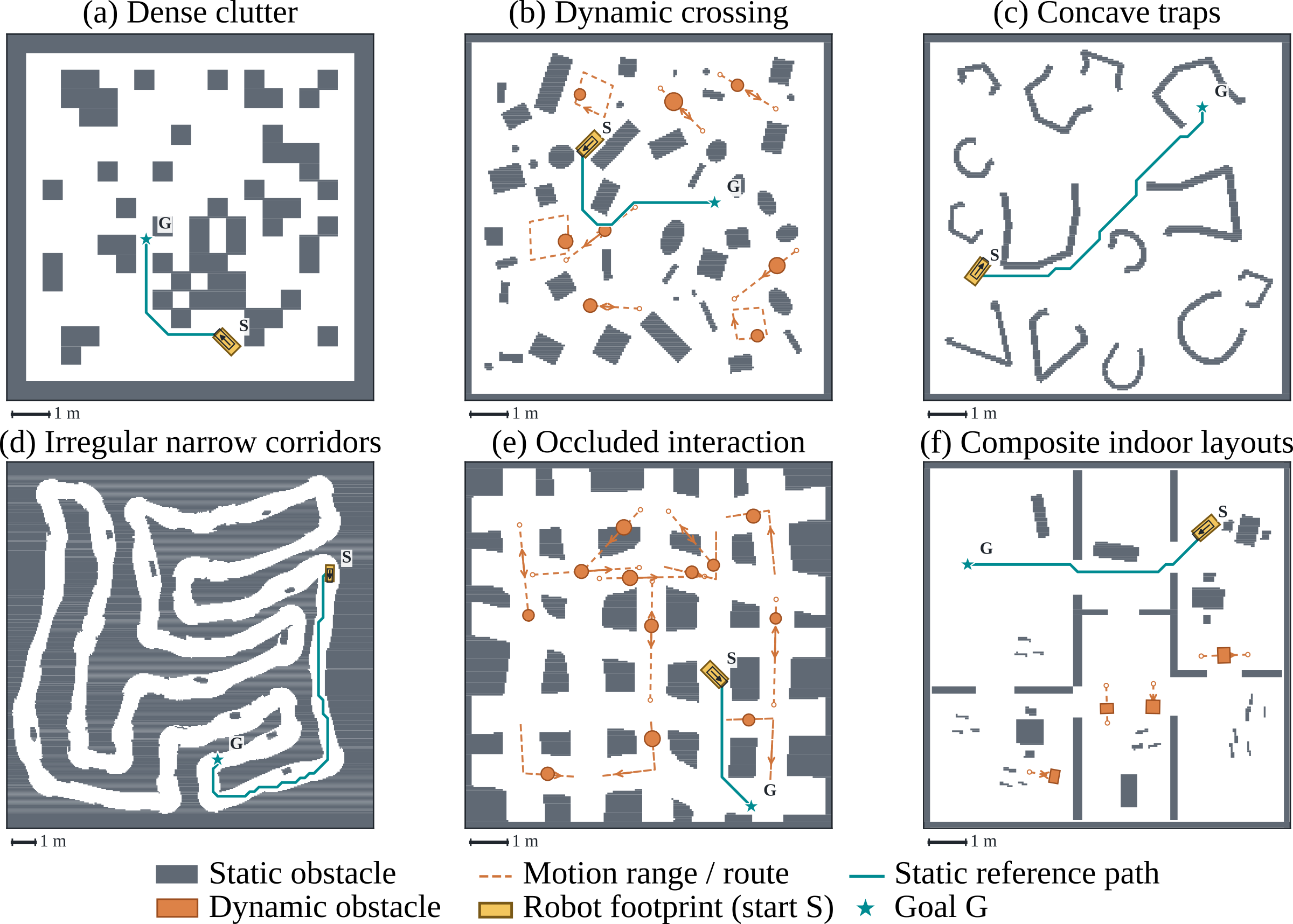}
    \caption{Representative layouts from the six training scene
    families, each containing 100 distinct maps.}
    \label{fig:training_maps}
    \vspace{-15pt}
\end{figure}

The training set comprises 600 layouts, evenly distributed across
six scene families: dense clutter, multidirectional dynamic crossings,
concave traps, irregular narrow corridors, occluded dynamic interactions,
and composite indoor layouts (Fig.~\ref{fig:training_maps}).
Doorway traversal is incorporated into indoor layouts.
Procedural generation varies obstacle geometry, clearance, occlusion,
and dynamic motion to cover complementary navigation challenges.

Each map provides a prevalidated pool of start–goal configurations and initial headings, sampled at episode reset. Validation checks footprint clearance, static connectivity after obstacle inflation, and dynamic trajectories
against map boundaries, static structures, and protected endpoint regions.
Scene labels, global maps, and reference paths are excluded from Actor
inputs; static cost-to-go information is used only for training-time
reward shaping and value estimation. This design broadens scenario
coverage while preserving the deployed policy's local-observation
interface.





\subsection{Asymmetric Domain Randomization}

We apply deployment-informed sensing and execution
randomization~\cite{tobin2017domain}. The Student distills Teacher
latents from corrupted histories, while the Teacher and Critic
retain clean observations and privileged states. Execution
perturbations affect both branches; clean observations reflect
the realized robot state.

LiDAR operates at $10\,\mathrm{Hz}$ with $0$--$20\,\mathrm{ms}$
additional delay, $1$--$2\,\mathrm{cm}$ range-noise standard deviation,
and $0$--$3\%$ ray dropout. Whole-scan dropout is capped at $0.3\%$,
with the last valid scan retained. Extrinsic perturbations,
$0$--$40\,\mathrm{ms}$ odometry delays, and slow pose drift further
corrupt Student observations.

Axis-dependent execution models capture tracking dynamics and
acceleration limits. Forward-motion delay, response time constant,
and gain span $0$--$40\,\mathrm{ms}$, $0.08$--$0.16\,\mathrm{s}$,
and $0.80$--$0.95$, respectively, with a startup dead zone of
$0.22$--$0.28\,\mathrm{m/s}$ and nonlinear low-speed attenuation.
Previous-command observations store the final bounded and filtered
commands. System parameters are sampled per episode; sensor noise
and dropout are sampled at each sensor update.

\section{Experiments}

\begin{figure}[t]
    \centering
    \vspace{5pt}
    \includegraphics[width=\columnwidth]{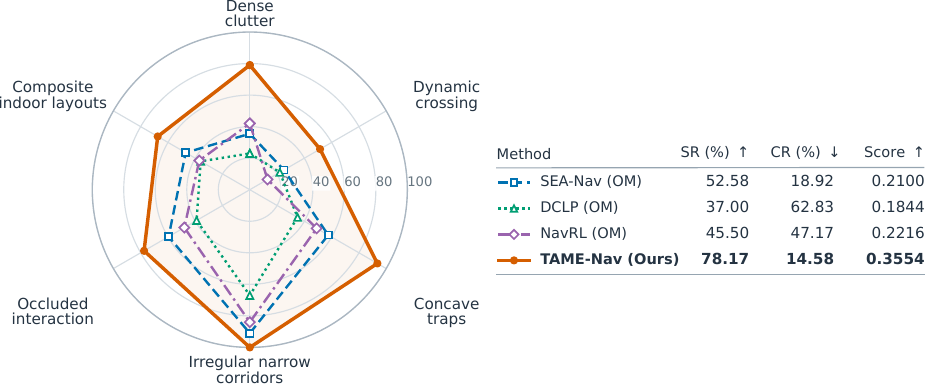}
    \caption{In-distribution navigation performance.
        (OM) denotes training on our maps. TAME-Nav achieves the highest success rate across all six scene
families and the best overall SR, CR, and Score, with particularly
clear gains in dense clutter and concave traps.}
    \label{fig:in_distribution_results}
    \vspace{-15pt}
\end{figure}

\subsection{Experimental Setup}

\subsubsection{Baselines}

We compare against three learning-based navigation methods.

\textbf{SEA-Nav}~\cite{chen2026sea} combines differentiable safety constraints and collision replay for efficient learning of safe, agile quadruped navigation in cluttered environments.

\textbf{NavRL}~\cite{xu2025navrl} combines reinforcement learning with velocity-obstacle-based safety for UAV navigation in dynamic environments. The authors also provide a Go2 quadruped deployment example in simulation.

\textbf{DCLP}~\cite{zhang2025drl} uses dimension-configurable representations and curriculum learning to accommodate varying robot dimensions and motion constraints without additional fine-tuning.

\subsubsection{Training}
We compare baselines trained on their original maps with methods
trained from scratch on our map set; the latter baseline variants
are denoted (OM). Baselines retain their original reward functions
and training configurations, with aligned interaction budgets.
TAME-Nav, SEA-Nav, and NavRL use equal numbers of environment
interaction steps, while SAC-based DRL-DCLP adjusts its step
budget according to the control period to match cumulative
simulated time. Each method is evaluated using its best validation
checkpoint within the allocated budget.



\subsubsection{Evaluation Metrics}

We report success rate (SR) and collision rate (CR). To jointly assess
task completion and efficiency, we adopt the BARN navigation
score used in DRL-DCLP~\cite{zhang2025drl}, as defined in
Eq.~\eqref{eq:navigation_score}:
\begin{equation}
\mathrm{Score}
=
\frac{1}{N}
\sum_{i=1}^{N}
S_i
\frac{T_i^\ast}
{\operatorname{clip}(T_i,\,2T_i^\ast,\,8T_i^\ast)},
\label{eq:navigation_score}
\end{equation}
where $N$ is the number of evaluation episodes, $S_i$ indicates
success, and $T_i$ is the actual navigation time.
The reference time is $T_i^\ast=L_i^\ast/v_{\max}$, where
$L_i^\ast$ is the shortest path length for the episode's
start--goal pair on the inflated static occupancy grid and
$v_{\max}=2.0\,\mathrm{m/s}$.
The clipping operator bounds the denominator between
$2T_i^\ast$ and $8T_i^\ast$. Failed episodes receive zero,
and higher scores indicate better overall performance.

\subsection{Simulation Experiments}

\subsubsection{In-Distribution Navigation Performance}
\label{sec:in_distribution}

We evaluate in-distribution generalization on 600 unseen layouts
independently generated from the same procedural scene distribution
as the training set.


\begin{figure}[!t]
    \centering
    \vspace{5pt}
    \includegraphics[width=0.6\linewidth]{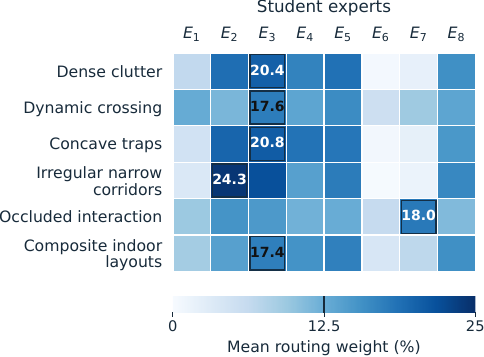}
    \caption{Mean expert routing weights across scene families.
    Boxed cells indicate the largest weight in each row;
    $12.5\%$ corresponds to uniform routing.}
    \label{fig:student_moe_routing}
    \vspace{-15pt}
\end{figure}

As shown in Fig.~\ref{fig:in_distribution_results}, TAME-Nav
achieves the highest SR across all six scene families.
The largest gains occur in dense clutter and concave traps,
exceeding 35 percentage points over the best baseline in each
family. By contrast, corridor navigation is already comparatively
strong for SEA-Nav (OM) and NavRL (OM).
Overall, SEA-Nav (OM) limits collisions but incurs more
non-collision failures, whereas collisions dominate the failures
of DCLP (OM) and NavRL (OM). TAME-Nav reduces both failure types,
and its leading Score indicates that improved task completion
is accompanied by stronger time-normalized performance.


We further examine expert routing across the six scene families
in Fig.~\ref{fig:student_moe_routing}. $E_3$ receives the largest
mean weight in dense clutter, dynamic crossings, concave traps,
and composite indoor layouts, with $E_5$ receiving nearly equal
weight in the latter. Narrow corridors favor $E_2$, whereas
occluded interactions assign the greatest weight to $E_7$.
Dense clutter and concave traps exhibit similar routing profiles,
indicating substantial expert sharing. These patterns suggest
that Soft-MoE adapts shared expert contributions to different
navigation demands through scene-dependent routing preferences,
supporting flexible specialization without a fixed correspondence
between scene families and experts.

\subsubsection{Sequential Compositional Navigation}

\begin{figure}[t]
    \centering
    \vspace{2pt}
    \includegraphics[width=0.8\linewidth]{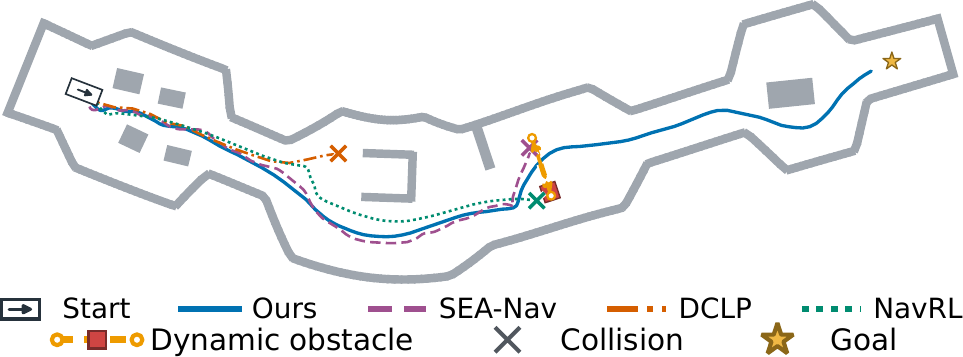}
    \caption{An example layout from the sequential compositional test set,
    with trajectories of our method and the OM baselines.}
    \label{fig:sequential_benchmark}
    \vspace{-10pt}
\end{figure}

We construct $64$ randomized test layouts, with one example shown in
Fig.~\ref{fig:sequential_benchmark}. Each layout combines dense clutter,
narrow doorways, dead-end pockets, occluded dynamic crossings, and
constrained goal approaches along a single channel without external
bypasses. Geometry and obstacle motion are randomized, and all layouts
pass reachability checks. The $32$ forward and $32$ reverse tasks,
each with eight matched dynamic-phase configurations, yield $512$
episodes. All methods use identical initial conditions and
deterministic mean actions.

We compare baselines trained on their original maps and ours.
For DRL-DCLP, whose complete training code and maps are not publicly
available, we report only our reproduction trained on our maps.

\begin{table}[t]
    \centering
    \caption{Sequential compositional navigation performance.
    Our method achieves the highest SR and Score and the lowest CR.
    (OM) denotes training on our maps; unmarked baselines use their
    original maps.}
    \label{tab:sequential_results}
    \small
    \setlength{\tabcolsep}{6pt}
    \begin{tabular}{lrrr}
        \toprule
        Method & SR (\%) $\uparrow$ & CR (\%) $\downarrow$
               & Score $\uparrow$ \\
        \midrule
        SEA-Nav & 47.27 & 52.73 & 0.1934 \\
        NavRL   & 0.00  & 75.39 & 0.0000 \\
        \midrule
        SEA-Nav (OM) & 66.80 & 33.01 & 0.2393 \\
        DCLP (OM)    & 59.96 & 40.04 & 0.2998 \\
        NavRL (OM)   & 34.96 & 60.94 & 0.1743 \\
        \textbf{TAME-Nav (Ours)}
                     & \textbf{81.64}
                     & \textbf{18.16}
                     & \textbf{0.4068} \\
        \bottomrule
    \end{tabular}
    \vspace{-15pt}
\end{table}


As shown in Table~\ref{tab:sequential_results}, TAME-Nav achieves
the highest SR and Score with the lowest CR. Collisions account
for nearly all failures of SEA-Nav (OM) and DCLP (OM), and dominate
those of NavRL (OM); the latter two incur more than twice our CR.
These results highlight the difficulty of maintaining safe progress
across successive challenges and support TAME-Nav's more reliable
completion of composite tasks. Original NavRL records no successes,
but its adapted aerial-to-ground sensing and control interfaces
limit conclusions about native-system performance.

\subsubsection{Compositional Multi-Room Navigation}

\begin{figure}[!t]
    \centering
    \vspace{2pt}
    \includegraphics[width=0.6\linewidth]{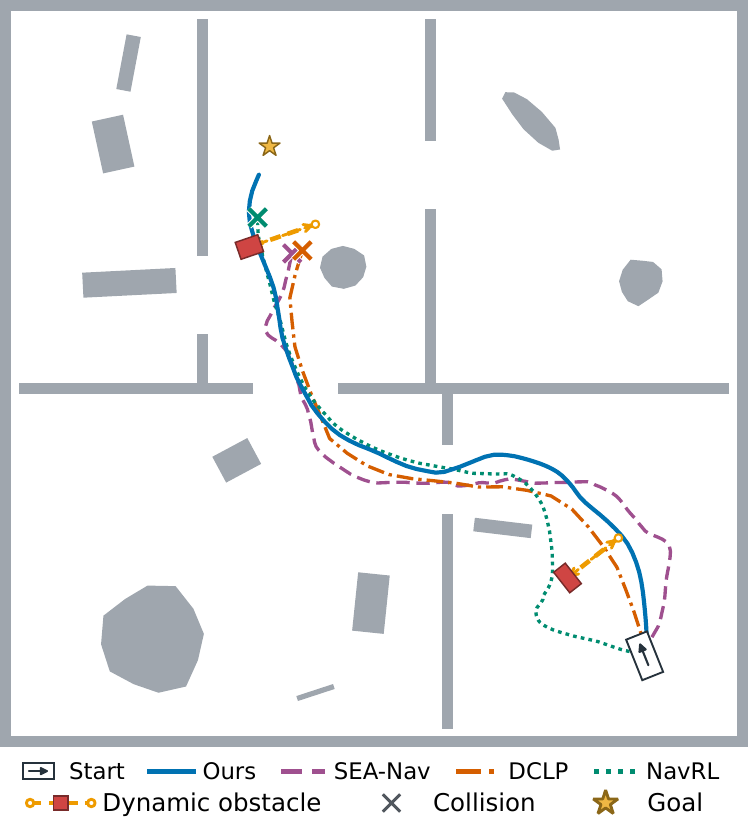}
    \caption{An example multi-room layout with paired trajectories of
    our method and the OM baselines. Dynamic obstacles are shown at their
    initial positions together with their motion ranges; each robot
    trajectory spans its full episode.}
    \label{fig:multiroom_benchmark}
    \vspace{-10pt}
\end{figure}

We construct a compositional multi-room test set comprising $64$ unseen
layouts, one of which is shown in Fig.~\ref{fig:multiroom_benchmark}.
Each layout contains $4$--$5$ connected rooms, doorways
$1.10$--$1.45\,\mathrm{m}$ wide, varied furniture, partially enclosed
structures, and $1$--$2$ dynamic obstacles. Room partitions, furniture
arrangements, start--goal pairs, and dynamic motion parameters are randomized,
and all layouts pass reachability and dynamic-trajectory validity checks.
Direct start-to-goal paths are obstructed, requiring policies to combine
cross-room traversal, occlusion handling, and dynamic obstacle avoidance.

We use the same models and paired evaluation protocol as in the preceding
experiment, with eight dynamic-phase configurations per map and $512$
episodes in total. All methods use identical start--goal pairs and dynamic
initial conditions, execute deterministic mean actions, and share a
$0.5\,\mathrm{m}$ goal threshold and a $60\,\mathrm{s}$ timeout.

\begin{table}[t]
    \centering
    \vspace{5pt}
    \captionsetup{font=small}
    \caption{Compositional multi-room navigation performance.
    Our method achieves the highest success rate and Score and the
    lowest collision rate in unseen multi-room environments.}
    \label{tab:multiroom_results}
    \small
    \setlength{\tabcolsep}{6pt}
    \begin{tabular}{lrrr}
        \toprule
        Method & SR (\%) $\uparrow$ & CR (\%) $\downarrow$
               & Score $\uparrow$ \\
        \midrule
        SEA-Nav & 48.24 & 37.70 & 0.2234 \\
        NavRL   & 3.91  & 42.97 & 0.0195 \\
        \midrule
        SEA-Nav (OM) & 66.41 & 22.85 & 0.2535 \\
        DCLP (OM)    & 60.16 & 35.16 & 0.2951 \\
        NavRL (OM)   & 67.58 & 15.62 & 0.3115 \\
        \textbf{TAME-Nav (Ours)}
                     & \textbf{75.20}
                     & \textbf{13.48}
                     & \textbf{0.3448} \\
        \bottomrule
    \end{tabular}
    \vspace{-15pt}
\end{table}

The multiroom results in Table~\ref{tab:multiroom_results} show
that TAME-Nav leads in SR and Score while maintaining the lowest
CR. NavRL (OM) remains the closest baseline, whereas DCLP (OM)
incurs more than twice our collision rate. The example trajectories
highlight failures during dynamic-obstacle encounters: all three
OM baselines collide, while TAME-Nav negotiates both interaction
regions and reaches the goal. This case illustrates how TAME-Nav
combines cross-room traversal with dynamic avoidance to sustain
progress toward the goal.

\subsection{Real-World Experiments}

We conduct real-world experiments on a Unitree Go2 equipped with
a Livox MID-360 LiDAR and an onboard NVIDIA Jetson Orin.
The high-level navigation policy issues velocity commands at
$50\,\mathrm{Hz}$, while LiDAR observations are updated at
$10\,\mathrm{Hz}$.

\subsubsection{Real-World Navigation in a Local-Trap Scenario}

\begin{figure}[t]
    \centering
    \vspace{2pt}
    \includegraphics[width=0.95\linewidth]{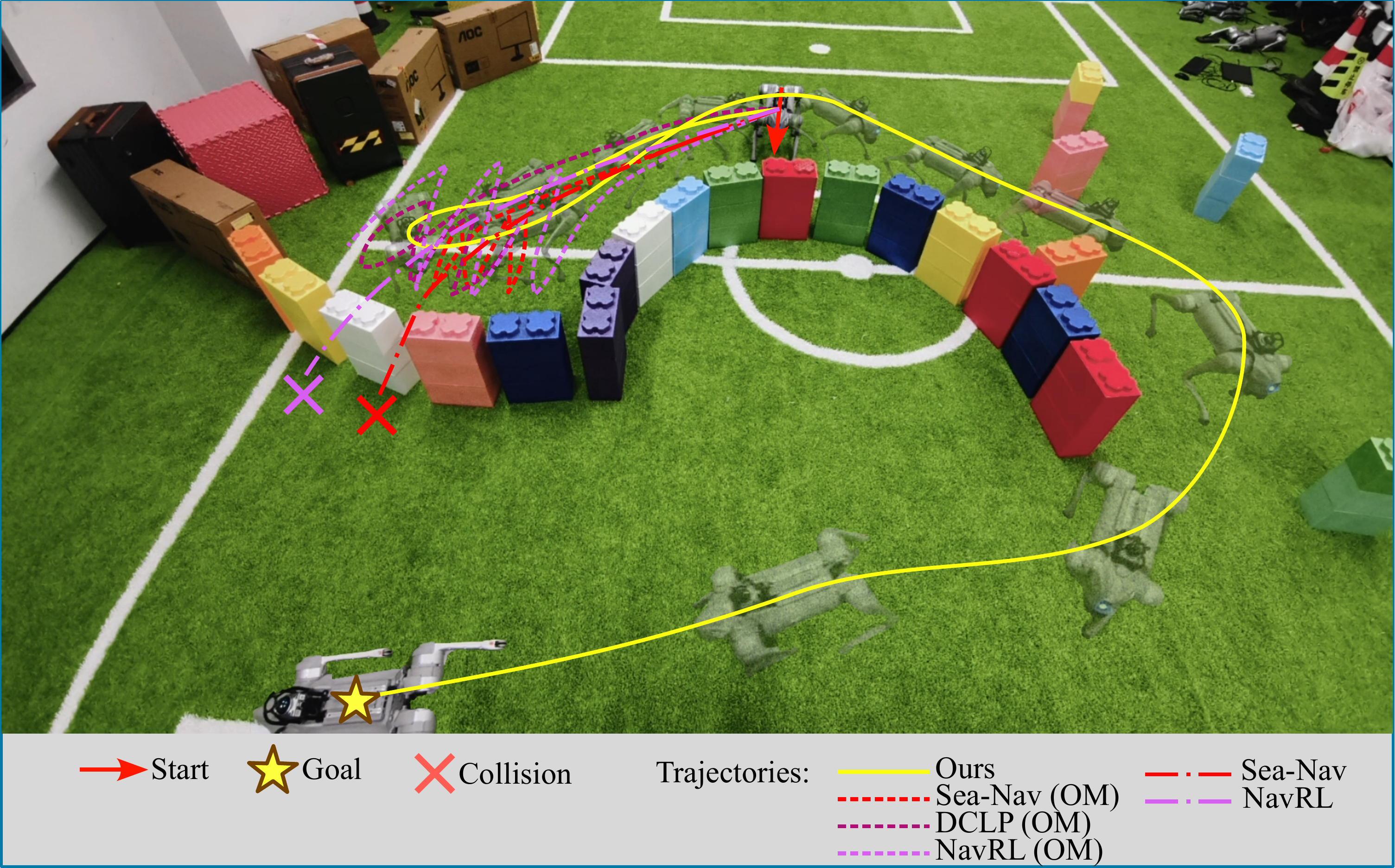}
    \caption{Real-world local-trap scenario and illustrative trajectories.
    Our method temporarily deviates from the goal direction to bypass the
    concave barrier and reach the goal.}
    \label{fig:real_escape}
    \vspace{-10pt}
\end{figure}

We construct a local-trap scenario comprising a concave barrier and surrounding
obstacles, as shown in Fig.~\ref{fig:real_escape}. The barrier blocks the direct
start-to-goal route, requiring the robot to temporarily deviate from the goal
direction and escape the trap. Each method is evaluated over ten trials.
A trial permits one minor collision that does not topple an obstacle; a second
collision, any collision that topples an obstacle, or persistent entrapment
without progress is counted as failure.

\begin{table}[t]
    \centering
    \vspace{5pt}
    \captionsetup{font=small}
    \caption{Real-world local-trap navigation over ten trials per method.
    Our method completes all trials with only one minor contact in total,
    whereas no baseline exceeds 50\% success.
    Collision counts are averaged per trial. One minor collision is allowed; a second collision, obstacle
toppling, or persistent entrapment constitutes failure.}
    \label{tab:real_escape}
    \small
    \setlength{\tabcolsep}{8pt}
    \begin{tabular}{lrr}
        \toprule
        Method & SR(\%)$\uparrow$ & Collisions/trial$\downarrow$ \\
        \midrule
        SEA-Nav       & 0   & 0.30 \\
        NavRL         & 0   & 1.00 \\
        \midrule
        SEA-Nav (OM)  & 0   & \textbf{0.00} \\
        DCLP (OM)     & 0   & 0.20 \\
        NavRL (OM)    & 50  & 0.80 \\
        \textbf{TAME-Nav (Ours)} & \textbf{100} & 0.10 \\
        \bottomrule
    \end{tabular}
    \vspace{-15pt}
\end{table}


As shown in Table~\ref{tab:real_escape}, our method completes all ten
trials with only one minor collision. NavRL (OM) achieves $50\%$ SR
with $0.8$ collisions per trial; some escapes occur when large-amplitude
oscillations reorient the robot toward a traversable direction.
All remaining baselines fail. Notably, SEA-Nav (OM) avoids collisions
but remains trapped, highlighting the distinction between immediate
collision avoidance and effective trap recovery. Despite training on
our multi-scenario maps, the OM baselines do not achieve reliable
trap escape in this real-world setting.

\subsubsection{Unexpected Dynamic Obstacles in Narrow Passages}

\begin{figure}[!t]
    \centering
    \vspace{2pt}
    \includegraphics[width=0.95\linewidth]{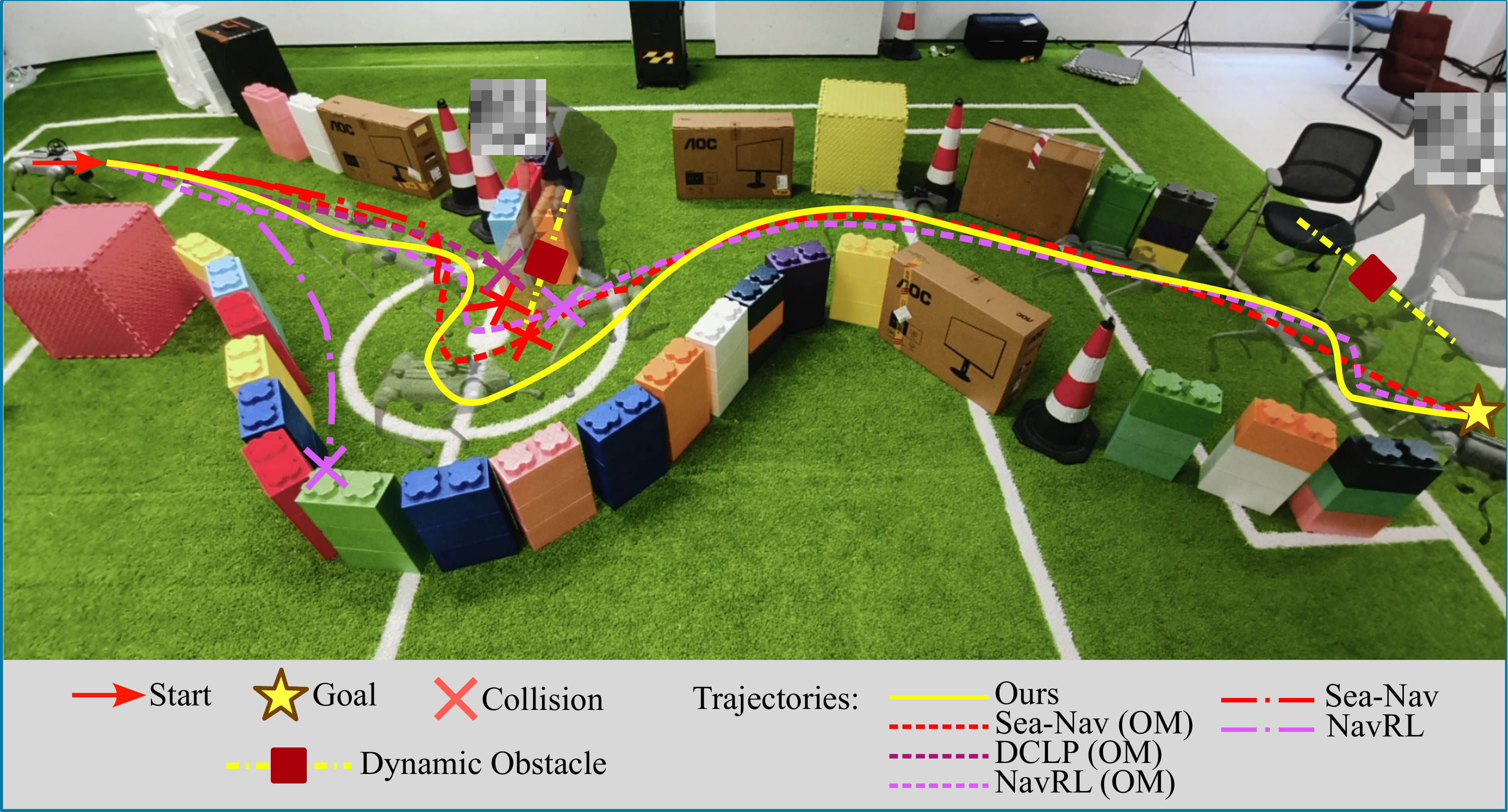}
    \caption{Narrow-passage scenario with unexpected dynamic obstacles
    and illustrative trajectories. Obstacles abruptly enter the robot's
    path midway through the passage and near the goal. Our method
    negotiates both encounters and reaches the goal.}
    \label{fig:real_dynamic}
    \vspace{-10pt}
\end{figure}

As shown in Fig.~\ref{fig:real_dynamic}, we construct a winding narrow
passage with dynamic obstacles abruptly entering the robot's path
midway and near the goal. The task requires passage following,
reactive avoidance, and goal approach within limited maneuvering
space. Each method is evaluated over ten trials using the same
collision and failure criteria as the preceding experiment.

\begin{table}[t]
    \centering
    \vspace{5pt}
    \caption{Real-world dynamic obstacle avoidance in narrow passages
    over ten trials per method. Our method completes all trials without
    collisions, while successful baselines incur repeated contacts.
    Collision counts are averaged per trial.}
    \label{tab:real_dynamic}
    \small
    \setlength{\tabcolsep}{8pt}
    \begin{tabular}{lrr}
        \toprule
        Method & SR(\%)$\uparrow$ & Collisions/trial$\downarrow$ \\
        \midrule
        SEA-Nav       & 80  & 1.00 \\
        NavRL         & 0   & 1.00 \\
        \midrule
        SEA-Nav (OM)  & 90  & 0.70 \\
        DCLP (OM)     & 80  & 0.70 \\
        NavRL (OM)    & 90  & 0.70 \\
        \textbf{TAME-Nav (Ours)} & \textbf{100} & \textbf{0.00} \\
        \bottomrule
    \end{tabular}
    \vspace{-15pt}
\end{table}


As shown in Table~\ref{tab:real_dynamic}, our method completes all ten
trials without collision. SEA-Nav (OM) and NavRL (OM) achieve
$90\%$ SR, versus $80\%$ for DCLP (OM); all three average $0.7$
collisions per trial. Original SEA-Nav achieves $80\%$ SR with
one collision per trial. Original NavRL records no successes,
although its aerial-to-ground transfer limits direct comparison.
This result demonstrates TAME-Nav’s ability to combine constrained-space traversal with collision-free avoidance of both unexpected dynamic obstacles.

\subsection{Ablation Study}


To assess the contributions of individual architectural components,
we compare the full model with three ablated variants. All variants
are evaluated using the same 600-map in-distribution test set and
evaluation protocol as in Sec.~\ref{sec:in_distribution}.

\begin{itemize}
\item \textbf{w/o MoE:}
Replaces the eight-expert Soft-MoE with a single parameter-matched MLP,
while retaining the same temporal encoder and latent dimension.

\item \textbf{w/o Teacher Tracks:}
Removes the Teacher's causal dynamic-track branch, retaining
its observation history.

\item \textbf{w/o Temporal History:}
Replaces the history with ten copies of the current observation,
preserving input dimensions and encoder architecture.

\end{itemize}


\begin{figure}[t]
    \centering
    \vspace{5pt}
    \includegraphics[width=\columnwidth]{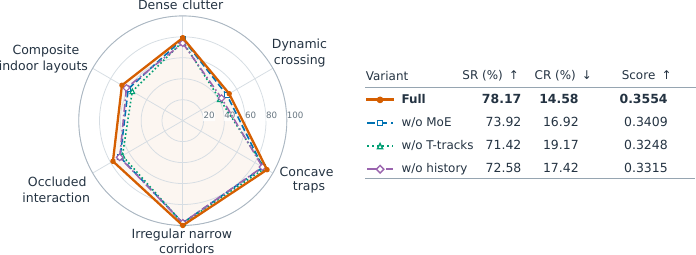}
    \caption{Ablation results on the in-distribution test set. All ablations degrade performance, with removing Teacher tracks causing the largest drop in success rate.}
    \label{fig:ablation}
    \vspace{-15pt}
\end{figure}


As shown in Fig.~\ref{fig:ablation}, the full model achieves the best
overall performance across the six scene families, with particularly
clear advantages in dynamic crossing, occluded interaction, and
composite indoor layouts, while all variants perform similarly in
irregular narrow corridors. The w/o MoE variant reduces SR by $4.25$
percentage points, while removing Teacher tracks or temporal history
causes larger drops of $6.75$ and $5.59$ points, respectively. These
results support the complementary roles of multi-expert representations,
privileged dynamic supervision, and temporal information.
\section{Conclusion}

We presented TAME-Nav, an asymmetric Teacher--Student framework
that distills track-augmented representations into a deployable
temporal Soft-MoE policy. Procedural scenario generation and
deployment-informed domain randomization support broad training
coverage and sim-to-real transfer. Benchmark evaluations demonstrate
improved navigation success, safety, and generalization, while
real-quadruped experiments validate local-trap escape and
narrow-passage traversal with sudden dynamic obstacles.

\bibliographystyle{IEEEtran}
\bibliography{reference}

@article{fox1997dynamic,
  title={The dynamic window approach to collision avoidance},
  author={Fox, Dieter and Burgard, Wolfram and Thrun, Sebastian},
  journal={IEEE robotics \& automation magazine},
  volume={4},
  number={1},
  pages={23--33},
  year={1997},
  publisher={IEEE}
}

@inproceedings{koren1991potential,
  title={Potential field methods and their inherent limitations for mobile robot navigation.},
  author={Koren, Yoram and Borenstein, Johann and others},
  booktitle={Icra},
  volume={2},
  pages={1398--1404},
  year={1991}
}

@inproceedings{pfeiffer2017perception,
  title={From perception to decision: A data-driven approach to end-to-end motion planning for autonomous ground robots},
  author={Pfeiffer, Mark and Schaeuble, Michael and Nieto, Juan and Siegwart, Roland and Cadena, Cesar},
  booktitle={2017 ieee international conference on robotics and automation (icra)},
  pages={1527--1533},
  year={2017},
  organization={IEEE}
}

@article{xu2022learning,
  title={Learning real-world autonomous navigation by self-supervised environment synthesis},
  author={Xu, Zifan and Nair, Anirudh and Xiao, Xuesu and Stone, Peter},
  journal={arXiv preprint arXiv:2210.04852},
  year={2022}
}

@inproceedings{rosmann2013efficient,
  title={Efficient trajectory optimization using a sparse model},
  author={R{\"o}smann, Christoph and Feiten, Wendelin and W{\"o}sch, Thomas and Hoffmann, Frank and Bertram, Torsten},
  booktitle={2013 European conference on mobile robots},
  pages={138--143},
  year={2013},
  organization={IEEE}
}

@inproceedings{tai2017virtual,
  title={Virtual-to-real deep reinforcement learning: Continuous control of mobile robots for mapless navigation},
  author={Tai, Lei and Paolo, Giuseppe and Liu, Ming},
  booktitle={2017 IEEE/RSJ international conference on intelligent robots and systems (IROS)},
  pages={31--36},
  year={2017},
  organization={IEEE}
}

@article{xie2023drl,
  title={Drl-vo: Learning to navigate through crowded dynamic scenes using velocity obstacles},
  author={Xie, Zhanteng and Dames, Philip},
  journal={IEEE Transactions on Robotics},
  volume={39},
  number={4},
  pages={2700--2719},
  year={2023},
  publisher={IEEE}
}

@article{hoeller2021learning,
  title={Learning a state representation and navigation in cluttered and dynamic environments},
  author={Hoeller, David and Wellhausen, Lorenz and Farshidian, Farbod and Hutter, Marco},
  journal={IEEE Robotics and Automation Letters},
  volume={6},
  number={3},
  pages={5081--5088},
  year={2021},
  publisher={IEEE}
}

@article{he2024agile,
  title={Agile but safe: Learning collision-free high-speed legged locomotion},
  author={He, Tairan and Zhang, Chong and Xiao, Wenli and He, Guanqi and Liu, Changliu and Shi, Guanya},
  journal={arXiv preprint arXiv:2401.17583},
  year={2024}
}

@article{ren2024top,
  title={Top-nav: Legged navigation integrating terrain, obstacle and proprioception estimation},
  author={Ren, Junli and Liu, Yikai and Dai, Yingru and Long, Junfeng and Wang, Guijin},
  journal={arXiv preprint arXiv:2404.15256},
  year={2024}
}

@article{pinto2017asymmetric,
  title={Asymmetric actor critic for image-based robot learning},
  author={Pinto, Lerrel and Andrychowicz, Marcin and Welinder, Peter and Zaremba, Wojciech and Abbeel, Pieter},
  journal={arXiv preprint arXiv:1710.06542},
  year={2017}
}

@inproceedings{chen2020learning,
  title={Learning by cheating},
  author={Chen, Dian and Zhou, Brady and Koltun, Vladlen and Kr{\"a}henb{\"u}hl, Philipp},
  booktitle={Conference on robot learning},
  pages={66--75},
  year={2020},
  organization={PMLR}
}

@article{lee2020learning,
  title={Learning quadrupedal locomotion over challenging terrain},
  author={Lee, Joonho and Hwangbo, Jemin and Wellhausen, Lorenz and Koltun, Vladlen and Hutter, Marco},
  journal={Science robotics},
  volume={5},
  number={47},
  pages={eabc5986},
  year={2020},
  publisher={American Association for the Advancement of Science}
}

@article{miki2022learning,
  title={Learning robust perceptive locomotion for quadrupedal robots in the wild},
  author={Miki, Takahiro and Lee, Joonho and Hwangbo, Jemin and Wellhausen, Lorenz and Koltun, Vladlen and Hutter, Marco},
  journal={Science robotics},
  volume={7},
  number={62},
  pages={eabk2822},
  year={2022},
  publisher={American Association for the Advancement of Science}
}

@inproceedings{monaci2022dipcan,
  title={DiPCAN: Distilling Privileged Information for Crowd-Aware Navigation.},
  author={Monaci, Gianluca and Aractingi, Michel and Silander, Tomi},
  booktitle={Robotics: Science and Systems},
  year={2022}
}

@article{zheng2026gpts,
  title={GPTS-Nav: End-to-End Robot Navigation in Dynamic Environments with Graph-Privileged Teacher-Student Reinforcement Learning},
  author={Zheng, Kairao and Li, Zhi and Yuan, Yiqing and Cheng, Hui},
  journal={IEEE Robotics and Automation Letters},
  year={2026},
  publisher={IEEE}
}

@inproceedings{messikommer2025student,
  title={Student-informed teacher training},
  author={Messikommer, Nico and Xing, Jiaxu and Aljalbout, Elie and Scaramuzza, Davide},
  booktitle={International Conference on Learning Representations},
  volume={2025},
  pages={67308--67323},
  year={2025}
}

@article{chen2026sea,
  title={SEA-Nav: Efficient Policy Learning for Safe and Agile Quadruped Navigation in Cluttered Environments},
  author={Chen, Shiyi and Yang, Mingye and Mao, Haiyan and Zhang, Jiaqi and Liu, Haiyi and He, Shuheng and Zhang, Debing and Qiu, Zihao and Zhang, Chun},
  journal={arXiv preprint arXiv:2603.09460},
  year={2026}
}

@article{xu2025navrl,
  title={NavRL: Learning safe flight in dynamic environments},
  author={Xu, Zhefan and Han, Xinming and Shen, Haoyu and Jin, Hanyu and Shimada, Kenji},
  journal={IEEE Robotics and Automation Letters},
  volume={10},
  number={4},
  pages={3668--3675},
  year={2025},
  publisher={IEEE}
}

@article{zhang2025drl,
  title={Drl-dclp: A deep reinforcement learning-based dimension-configurable local planner for robot navigation},
  author={Zhang, Wei and Wang, Shanze and Tan, Mingao and Yang, Zhibo and Wang, Xianghui and Shen, Xiaoyu},
  journal={IEEE Robotics and Automation Letters},
  volume={10},
  number={4},
  pages={3636--3643},
  year={2025},
  publisher={IEEE}
}

@article{cheng2024navila,
  title={Navila: Legged robot vision-language-action model for navigation},
  author={Cheng, An-Chieh and Ji, Yandong and Yang, Zhaojing and Gongye, Zaitian and Zou, Xueyan and Kautz, Jan and B{\i}y{\i}k, Erdem and Yin, Hongxu and Liu, Sifei and Wang, Xiaolong},
  journal={arXiv preprint arXiv:2412.04453},
  year={2024}
}

@article{zhang2026saferpath,
  title={SaferPath: Hierarchical Visual Navigation with Learned Guidance and Safety-Constrained Control},
  author={Zhang, Lingjie and Jiang, Zeyu and Chen, Changhao},
  journal={arXiv preprint arXiv:2603.01898},
  year={2026}
}

@inproceedings{trautman2010unfreezing,
  title={Unfreezing the robot: Navigation in dense, interacting crowds},
  author={Trautman, Peter and Krause, Andreas},
  booktitle={2010 IEEE/RSJ International Conference on Intelligent Robots and Systems},
  pages={797--803},
  year={2010},
  organization={IEEE}
}

@article{kaelbling1998planning,
  title={Planning and acting in partially observable stochastic domains},
  author={Kaelbling, Leslie Pack and Littman, Michael L and Cassandra, Anthony R},
  journal={Artificial intelligence},
  volume={101},
  number={1-2},
  pages={99--134},
  year={1998},
  publisher={Elsevier}
}

@article{schulman2017proximal,
  title={Proximal policy optimization algorithms},
  author={Schulman, John and Wolski, Filip and Dhariwal, Prafulla and Radford, Alec and Klimov, Oleg},
  journal={arXiv preprint arXiv:1707.06347},
  year={2017}
}

@article{vaswani2017attention,
  title={Attention is all you need},
  author={Vaswani, Ashish and Shazeer, Noam and Parmar, Niki and Uszkoreit, Jakob and Jones, Llion and Gomez, Aidan N and Kaiser, {\L}ukasz and Polosukhin, Illia},
  journal={Advances in neural information processing systems},
  volume={30},
  year={2017}
}

@inproceedings{cho2014learning,
  title={Learning phrase representations using RNN encoder--decoder for statistical machine translation},
  author={Cho, Kyunghyun and Van Merri{\"e}nboer, Bart and Gul{\c{c}}ehre, {\c{C}}a{\u{g}}lar and Bahdanau, Dzmitry and Bougares, Fethi and Schwenk, Holger and Bengio, Yoshua},
  booktitle={Proceedings of the 2014 conference on empirical methods in natural language processing (EMNLP)},
  pages={1724--1734},
  year={2014}
}

@article{jacobs1991adaptive,
  title={Adaptive mixtures of local experts},
  author={Jacobs, Robert A and Jordan, Michael I and Nowlan, Steven J and Hinton, Geoffrey E},
  journal={Neural computation},
  volume={3},
  number={1},
  pages={79--87},
  year={1991},
  publisher={MIT Press}
}

@inproceedings{tobin2017domain,
  title={Domain randomization for transferring deep neural networks from simulation to the real world},
  author={Tobin, Josh and Fong, Rachel and Ray, Alex and Schneider, Jonas and Zaremba, Wojciech and Abbeel, Pieter},
  booktitle={2017 IEEE/RSJ international conference on intelligent robots and systems (IROS)},
  pages={23--30},
  year={2017},
  organization={IEEE}
}

@article{wu2026toward,
  title={Toward reliable sim-to-real predictability for moe-based robust quadrupedal locomotion},
  author={Wu, Tianyang and Guo, Hanwei and Wang, Yuhang and Yang, Junshu and Sui, Xinyang and Xie, Jiayi and Chen, Xingyu and Liu, Zeyang and Lan, Xuguang},
  journal={arXiv preprint arXiv:2602.00678},
  year={2026}
}

\end{document}